\documentclass[conference]{IEEEtran}
\IEEEoverridecommandlockouts
\usepackage{cite}
\usepackage{amsmath,amssymb,amsfonts}
\usepackage{algorithmic}
\usepackage{graphicx}
\usepackage{textcomp}
\usepackage{xcolor}
\usepackage{pdfpages}

\def\BibTeX{{\rm B\kern-.05em{\sc i\kern-.025em b}\kern-.08em
    T\kern-.1667em\lower.7ex\hbox{E}\kern-.125emX}}
\begin{document}

\title{Rapid Autonomous Car Control based on Spatial and Temporal Visual Cues\\
}

\author{\IEEEauthorblockN{1\textsuperscript{st} Surya Dantuluri}
\IEEEauthorblockA{\textit{Monta Vista High School} \\
Cupertino, United States of America \\
dsuryav@gmail.com}
}

\maketitle

\begin{abstract}
We present a novel approach to modern car control utilizing a combination of Deep Convolutional Neural Networks and Long Short-Term Memory Systems:  Both of which are a subsection of Hierarchical Representations Learning, more commonly known as Deep Learning. Using Deep Convolutional Neural Networks and Long Short-Term Memory Systems (DCNN/LSTM), we propose an end-to-end approach to accurately predict steering angles and throttle values. We use this algorithm on our latest robot, El Toro Grande 1 (ETG) which is equipped with a variety of sensors in order to localize itself in its environment. Using previous training data and the data that it collects during circuit and drag races, it predicts throttle and steering angles in order to stay on path and avoid colliding into other robots. This allows ETG to theoretically race on any track with sufficient training data.
\end{abstract}


\section{Introduction}
Monta Vista High School's El Toro Grande 1 is a proof of concept, answering the question on whether Machine Learning methods can be applied to autonomous driving to rival traditional computationally expensive computer vision, path planning, and localization algorithms. This report explains these methods as well as the hardware innovations necessary to execute such new software methods. This report is an example of how the International Autonomous Robot Racing Challenge (IARRC) promotes advancement in research of autonomous vehicles at the secondary and university school levels.
\newline
\indent \textit{Vehicle Name:} Monta Vista High School's Robotics Team has a tradition of naming their robots as \textit{El Toro} (English translation: The Bull). Most of \textit{El Toro} robots are used for For Inspiration and Recognition of Science and Technology (FIRST) robotics competitions, held primarily in the USA. Every \textit{El Toro} robot is hardcoded to perform tasks during an autonomous period during FIRST robotics competitions and are manually controlled for the rest of the match. Our robot for the IARRC competition is relatively nimble and a lot more autonomous when compared to other \textit{El Toro} robots. As a result of the tradition and how advanced this robot is in terms of hardware and software, the name \textit{El Toro Grande 1} (English translation: The Big Bull) is a fitting for this robot.

\section{Platform Design}

\begin{figure}[ht]
\centering
\centerline{\includegraphics[width=6cm]{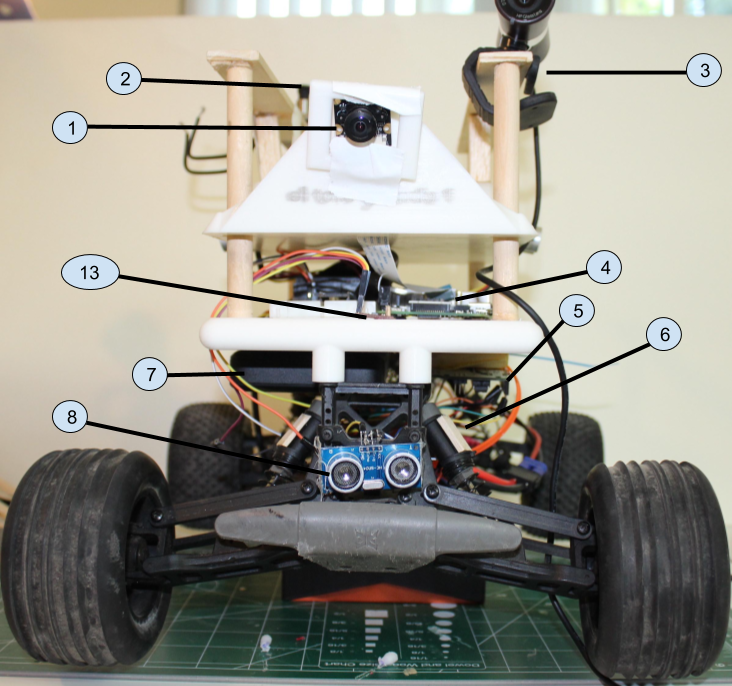}}
\caption{Front view of robot components. Cross-referenced in Tables 1,2, and 3}
\label{fig:front}
\end{figure}
\begin{figure}[ht]
\centerline{\includegraphics[width=7cm]{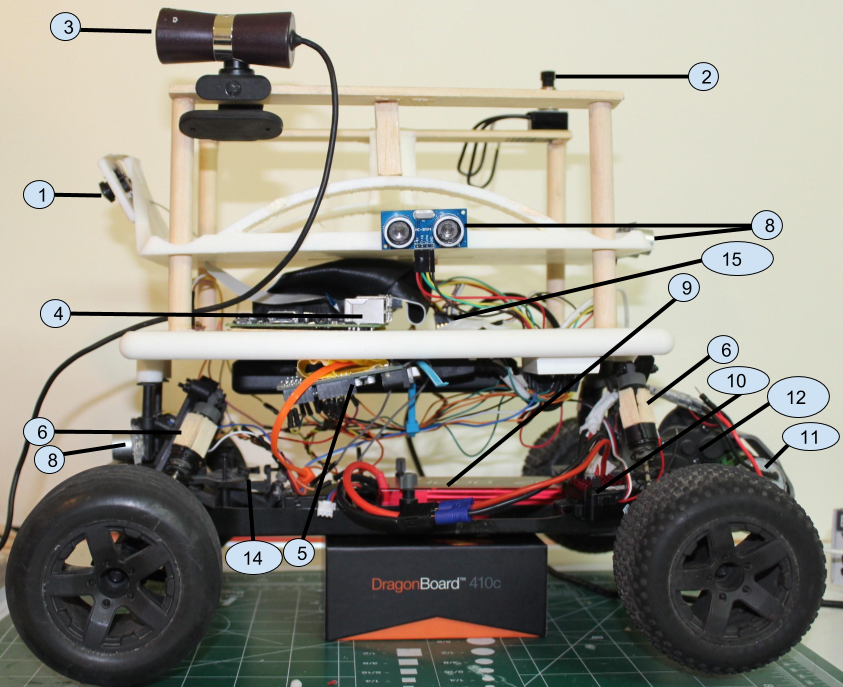}}
\caption{Side view of robot components. Cross-referenced in Tables 1,2, and 3}
\label{fig:side}
\end{figure}

\subsection{Platform Overview}
ETG is made up of 3 networks: Sensor Network, Vision Network, and the Engine Network.
Among all these networks, the Quad Core 1.2GHz Raspberry Pi 3 conducts computational flows among all three concurrently. The upper structure of ETG is 3D printed from a CAD model.
\subsection{Sensor Network}
\begin{figure}[ht]
\centerline{\includegraphics[width=7cm]{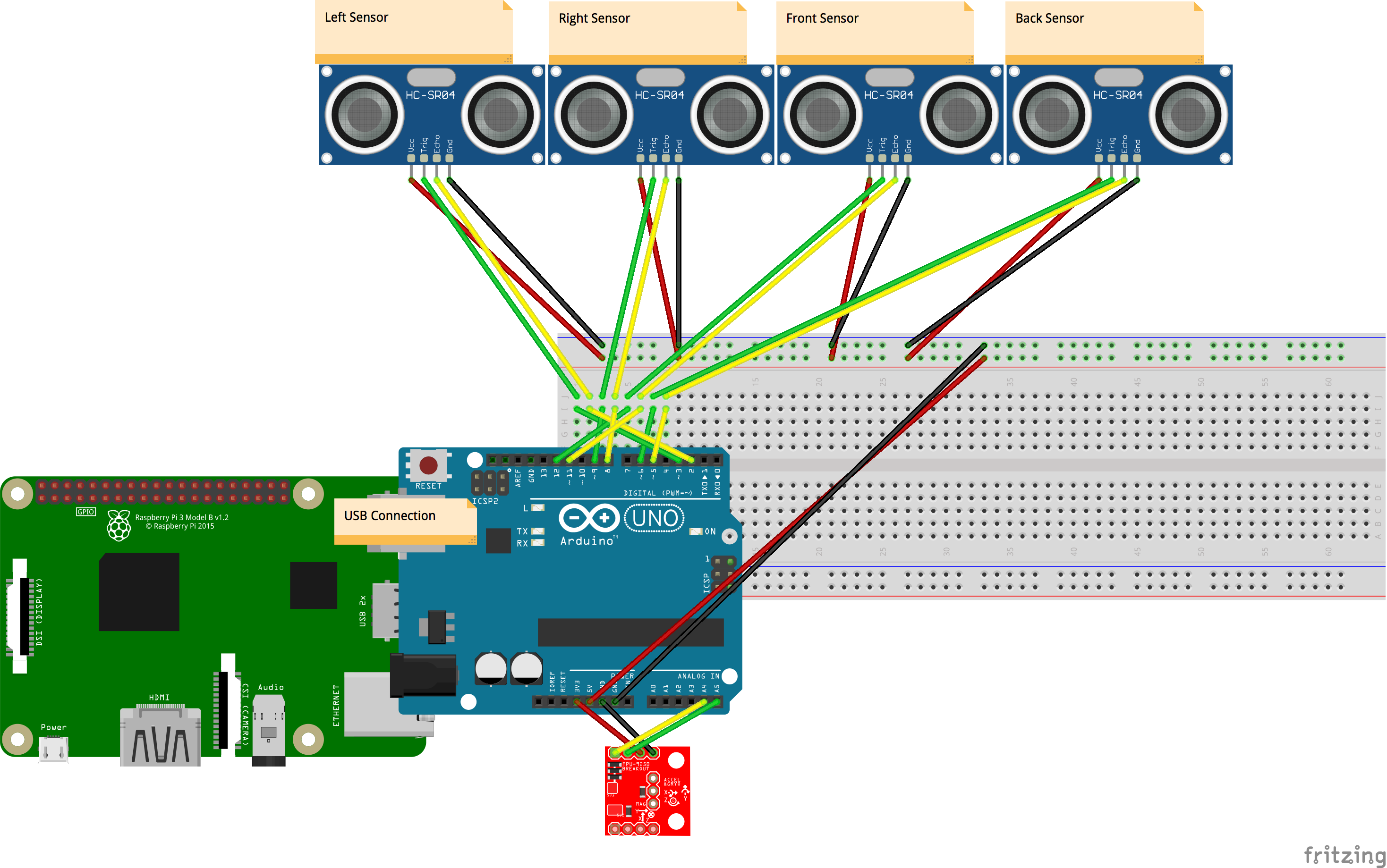}}
\caption{Sensor Network}
\label{fig:sensors}
\end{figure}

There are five sensors mounted around ETG. Four of these sensors are ultrasonic sensors placed on all four sides of ETG. All the ultrasonic sensors are used digitally and use a threshold to indicate whether the ultrasonic sensor will record data or not. This means that the sensors indicate something is present 50cm away with recording "1" in the data. If something is more than 50cm away from any of the ultrasonic sensors, the ultrasonic sensors will not record a "1" in the data, rather, a "0". The threshold has been put in place as a consequence of the noisiness of the HC-SR04 Ultrasonic Sensors. A SparkFun MPU-9250 IMU Breakout board is used to collect acceleration data as indicated in figure \ref{fig:sensors}

\begin{table}[htbp]
\caption{Sensor Network}
\begin{center}
\hskip-1.0cm
 \begin{tabular}{||c c c c||} 
 \hline
 \textbf{\#} & \textbf{Sensor} & \textbf{Type} & \textbf{Function} \\ [0.5ex]
 \hline\hline
 8 & Ultrasonic Sensor - HC-SR04 & Distance & Localization \\ 
 \hline
 13 & MPU-9250 IMU & Positioning & Localization \\ [1ex] 
 \hline
\end{tabular}
\end{center}
\end{table}

\subsection{Vision Network}
\begin{figure}[ht]
\centerline{\includegraphics[width=5cm]{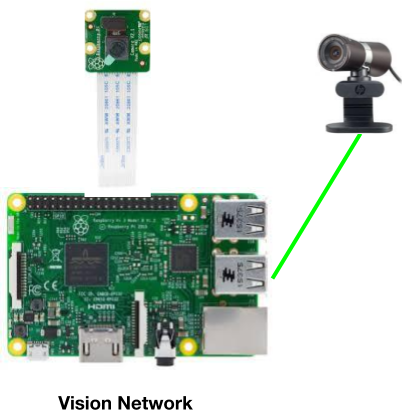}}
\caption{Vision Network}
\label{fig:vision}
\end{figure}

There are two crucial components to the Vision Network, the Wide-Angle Raspberry Pi Camera and the HP 4110 Webcam as shown in figure \ref{fig:vision}. The Wide-Angle Raspberry Pi Camera (Pi Camera) occupies most of the weight while training on DCNN/LSTM while the HP Webcam provides a stream of images of the traffic light to a trained Convolutional Neural Network model. Both cameras supply images to the Raspberry Pi 3 for throttle values and steering angle prediction.

\begin{table}[htbp]
\caption{Vision Network}
\begin{center}
\hskip1.0cm
 \begin{tabular}{||c c c c||} 
 \hline
 \textbf{\#} & \textbf{Camera} & \textbf{Type} & \textbf{Function} \\ [0.5ex]
 \hline\hline
 1 &  Wide Angle Fish-Eye Pi Camera & Camera & Track Detection \\ 
 \hline
 3 & HP 4110 Webcam & Camera & Traffic Light Detection \\ [1ex] 
 \hline
\end{tabular}
\end{center}
\end{table}

\subsection{Engine Network}
\begin{figure}[ht]
\centerline{\includegraphics[width=5cm]{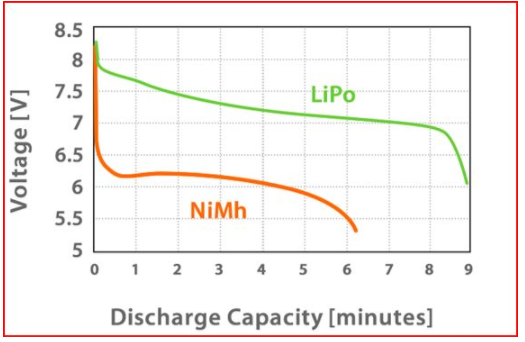}}
\caption{LiPo and NiMH voltage dropoff curves }
\label{fig:battery}
\end{figure}

The Engine Network has been completely refreshed from the stock configuration given in the original RC Car. This new network is equipped with a 27 turn motor, which significantly increases torque when compared to the stock 15 turn motor. A 4 Amp LiPo battery is used, rather than the stock 1.8 Amp NiMH battery because of the voltage drop off as shown in figure \ref{fig:battery}. Using a NiMH battery, throttle PWM values are inconsistent because PWM values tend to be higher at the end of the battery cycle in order to maintain the same velocity, whereas using LiPo batteries, throttle PWM values stay relatively the same throughout the battery charge cycle to maintain a constant velocity. Having a consistent throttle value cleans the data so that the LSTM will not have to find out this vague pattern in the data. A 60A ESC and a 5V BEC are used to receive PWM signals from the PCA 9685 and control the motor. A HITEC HS-645MG Servo has been mounted on the car to increase steering range and to increase $K_s$ in the Ackerman steering geometry equation as shown in equation \ref{eq:1}. By increasing $K_s$, ETG is capable of making harder turns left or right on difficult corners of the race. The Ackerman steering angle equation is described by Kim [3] as follows; $\theta_t$ is steering angle in degrees, $v_t$ (m/s) is the velocity at time t. $K_s$ is the steering ratio between the turn of the steering and the turn of the wheels, as described before. $K_{slip}$ represents the relative motion of the steering wheel with respect to the road and $d_w$ is the length between the front and rear wheels.

\begin{equation} \label{eq:1}
\theta_t = f_{steers} = (u_t)d_wK_s(1+K_{slip}v_t^2)
\end{equation}

\begin{figure}[ht]
\centerline{\includegraphics[width=7cm]{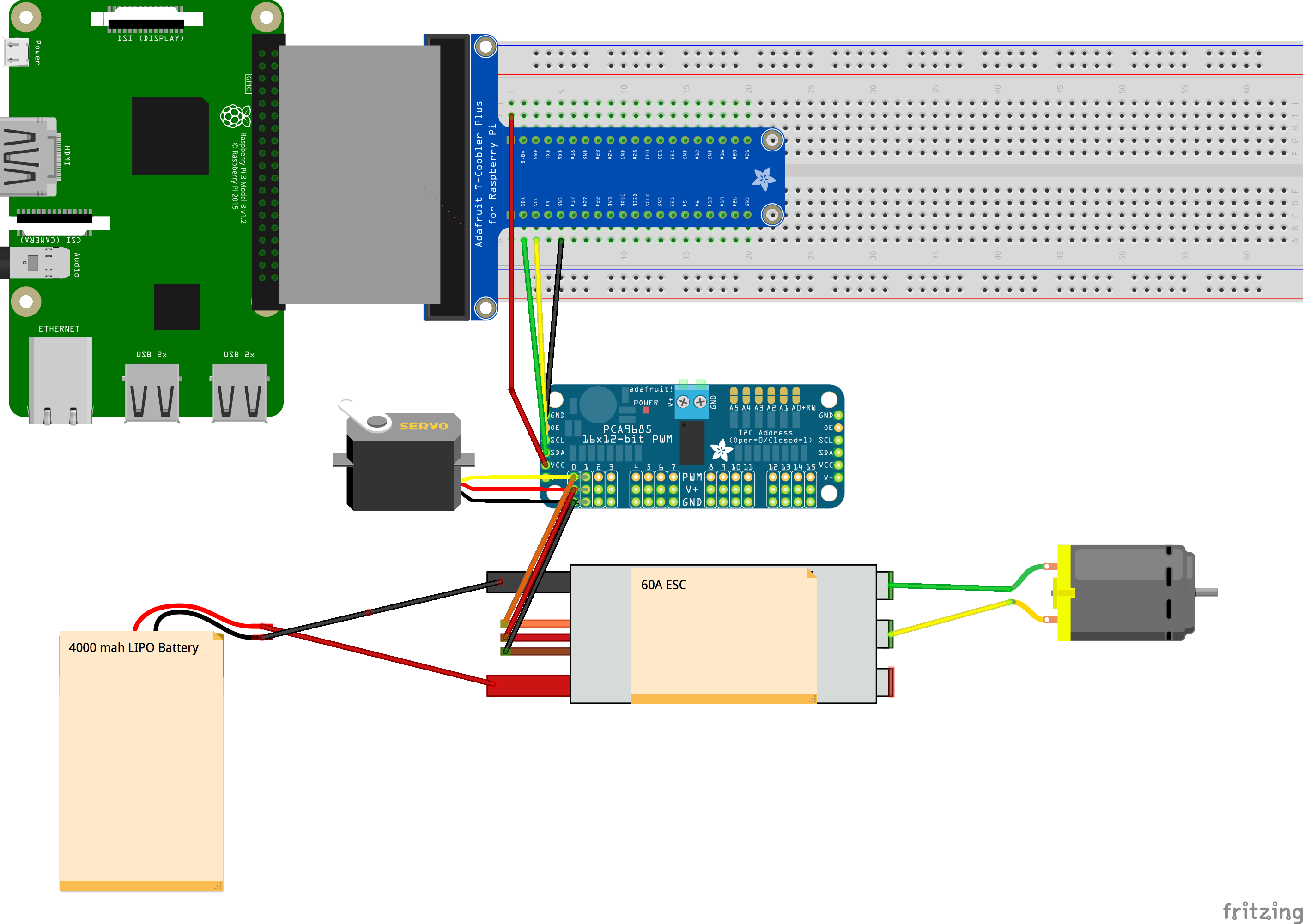}}
\caption{Engine Network}
\label{fig:motors}
\end{figure}

\begin{table}[htbp]
\caption{Engine Network}
\begin{center}
\hskip-1.0cm
 \begin{tabular}{||c c c c||} 
 \hline
 \textbf{\#} & \textbf{Part} & \textbf{Type} & \textbf{Function} \\ [0.5ex]
 \hline\hline
 10 &  60A Dynamite ESC & ESC & Motor Control \\ 
 \hline
 14 & HITEC HS-645MG  & Servo Motor & Traffic Light Detection \\ 
 \hline
 9 & 4000 MAH LiPo  & Battery & Power supply \\
 \hline
 15 & PCA9685  & PWM Control & PWM Output \\ 
  \hline
 11 & 27T Brushed Motor  & Brushed Motor & Throttle \\ [1ex] 
 \hline
\end{tabular}
\end{center}
\end{table}

\subsection{More Electro-Mechanical Engineering}
\begin{figure}[ht]
\centering
\centerline{\includegraphics[width=6cm]{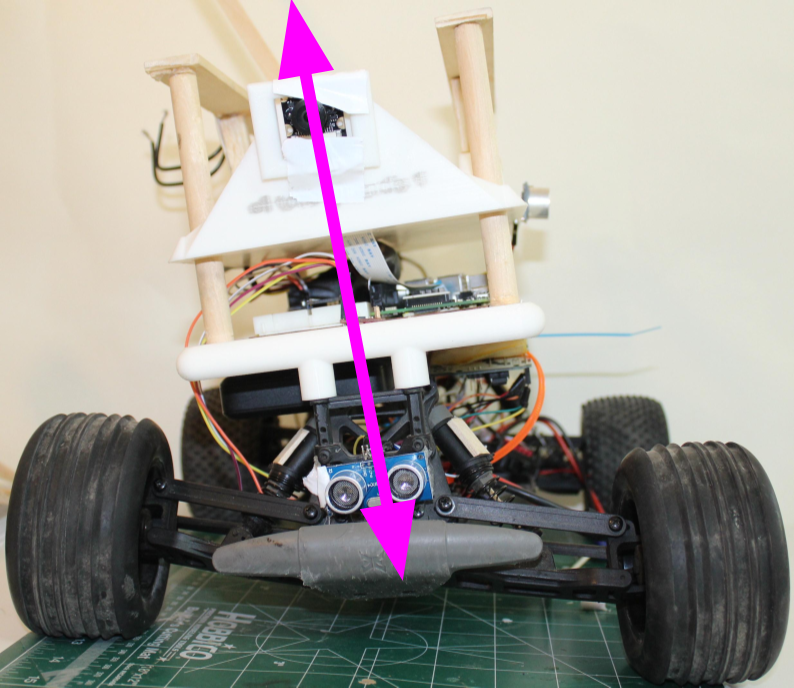}}
\caption{Line representing the normal force vector, which points roughly north-western, and the gravitational component antiparallel to the normal force vector}

\label{fig:left}
\end{figure}

\begin{figure}[ht]
\centering
\centerline{\includegraphics[width=6cm]{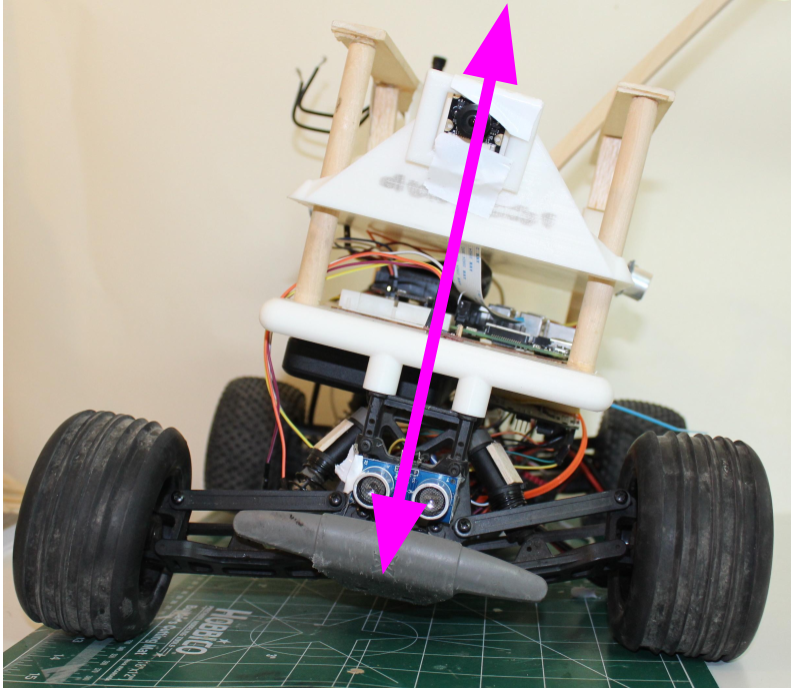}}
\caption{Line representing  the normal force vector, which points roughly north-eastern, and the gravitational component antiparallel to the normal force vector}
\label{fig:right}
\end{figure}

The stock ECX 1/10 Circuit 2WD has been heavily modified as described in the Engine Network subsection. Some modifications to the chassis are: 
\begin{itemize}
  \item Damped suspension
  \begin{itemize}
      \item Stock suspension allowed only 0.5 kg of load on the vehicle. By placing 2 wooden spacers 2cm long adjacent to suspension coils across the vehicle, load was increased to an upwards of 5 kg.
      \item An increased load caused the vehicle to tilt left or right in order to maintain ground force. Simulated in figures \ref{fig:left} and \ref{fig:right}.
  \end{itemize}
  \item Increased tooth count for the pinion gear
  \begin{itemize}
      \item The stock 15 tooth pinion gear was changed for a 20 tooth pinion gear.
      \item A higher tooth count for the pinion gear increases speed but slows acceleration.
      \item A higher tooth count is necessary since the 27T motor made the robot noticeably slower from the stock 15T motor. This higher speed allowed the ETG to go at speeds up to 12 m/s.
  \end{itemize}
\end{itemize}
\subsection{Vehicle Body}
\begin{figure}
  \centering
  \hskip-1.0cm
  \includegraphics[width=9cm]{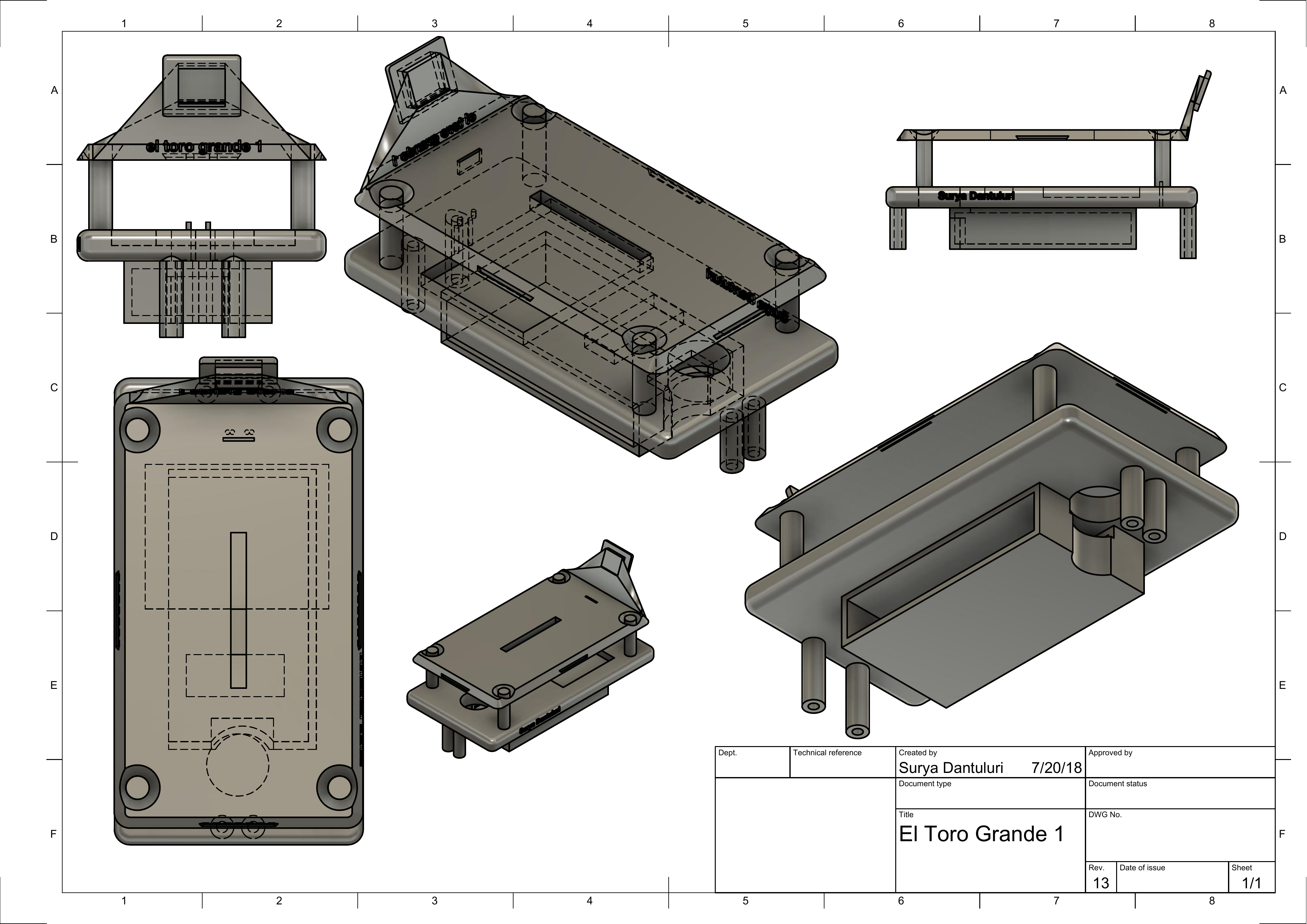}
  \caption{CAD Drawing for ETG upper body}
  \label{fig:cad}
\end{figure}

The vehicle body was created to be mounted on top of the chassis of the ECX 1/10 Circuit car. It was created using Fusion 360 and printed through 3D Hubs, an online 3D printing service. The parts were shipped from Ohio and were assembled with wooden dowels and Balsa wood. Using the wooden support system, 3D printing costs were reduced by an upwards of 10 times the original price of the CAD model. The original CAD model is shown in figure \ref{fig:cad}.


\section{Hardware Systems Design}
\subsection{Processor Framework}
Processing is distributed among the Raspberry Pi 3 (RPI3) and the Arduino UNO R3. During the data collection stage, the RPI3 collects sensor data from the Arduino and stores that sensor data with image data. This data is trained on the DCNN/LSTM on a cloud server to generate a model. This model is run on the RPI3, which uses sensor data and image data to make throttle and steering angle value predictions.
\subsection{Electrical Sub-systems}
A 10 amp 5v portable battery charger powers the RPI3. The RPI3 provides power to the Arduino through a USB port and powers the PCA 9685 to generate PWM signals to control the servo and ESC. A 4 amp 7.2v 2 cell LiPo battery powers the ESC which controls the 27T motor.

\section{Perception And Control}
\subsection{Software Architecture}
\begin{figure}[ht]
\centering
\centerline{\includegraphics[width=7cm]{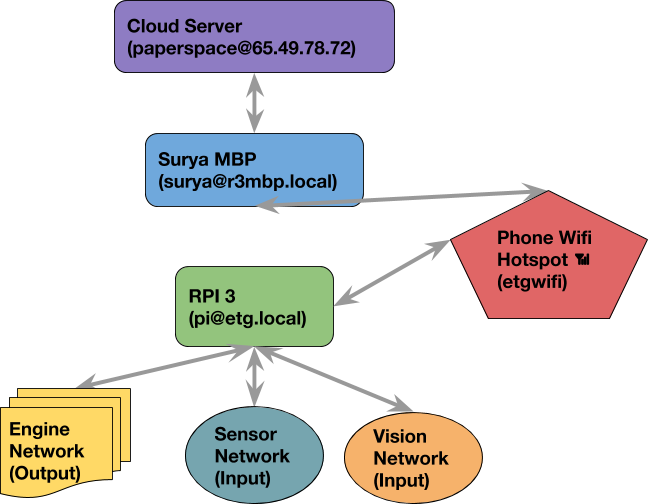}}
\caption{Software Architecture}
\label{fig:architecture}
\end{figure}

The RPI3 is the central processor, conducting all the subsystems to work together. The Sensor Network works as follows; The Arduino UNO is responsible for polling the 4 Ultrasonic sensors located on all four sides around the vehicle as well as the MPU 9250 IMU. Generally, with a required buffer included to effectively poll all sensors within a couple milliseconds, a steady rate of 12 readings per second is achieved. These readings are sent to the Raspberry Pi 3 through the UART serial port using a UDP connection as shown in figure \ref{fig:architecture}. The Raspberry Pi 3 only receives readings when it writes to the Arduino through the serial port, requesting a reading. This occurs every time the main program on the Raspberry Pi 3 goes through the drive loop(explained in the next section). The drive loop on the Raspberry Pi 3 occurs at a rate of around 20-30 times a second. As a result, the RPI3 and Arduino protocol are forced to be done on a separate thread. This allows the drive loop to run efficiently while also getting sensor readings at a steady rate. The RPI3 performs all the high-level processing, including data management and data retrieval. A cloud server preprocesses and runs the DCNN/LSTM model on the data collected by the RPI3. Generally, this takes 30-40 minutes to train on a dataset of 40,000 images. The cloud server consists of 8 CPU cores, 30 gigabytes of RAM, and a Nvidia Quadro P4000. The model uses Tensorflow GPU to train significantly faster. The trained model is then run on the RPI3, which uses all the networks to predict throttle values and steering angles. The Surya MBP node is used to start the car through SSH which is done through the Wifi hotspot as shown in figure \ref{fig:architecture}.
\subsubsection{Drive Loop}
The drive loop is a loop that polls every component in the Engine, Vision, and Sensor network. The drive loop is used for both the data collection stage and running stage. In the data collection stage, the loop polls components as follows: Pi Camera, 4 Ultrasonic Sensors and 1 IMU (run on a eparate thread), steering angle, and throttle value. This data is stored in JSON files with references to the corresponding image (since an image is taken for every iteration of the drive loop). During the running stage, the loop polls the same components except for the engine network. The vision and sensor network are used as parameters for the trained model which outputs throttle and steering angle value predictions.

\subsection{Perception and Planning}
\subsubsection{Object Detection}
\begin{figure}[ht]
\centering
\centerline{\includegraphics[width=5cm]{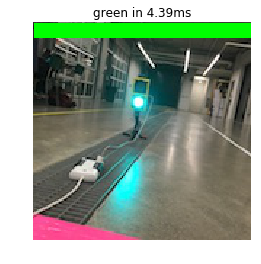}}
\caption{Green prediction}
\label{fig:green}
\end{figure}

\begin{figure}[ht]
\centering
\centerline{\includegraphics[width=5cm]{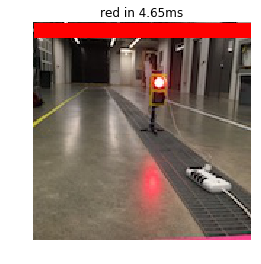}}
\caption{Red prediction}
\label{fig:red}
\end{figure}

\begin{figure}[ht]
\centering
\centerline{\includegraphics[height=16cm]{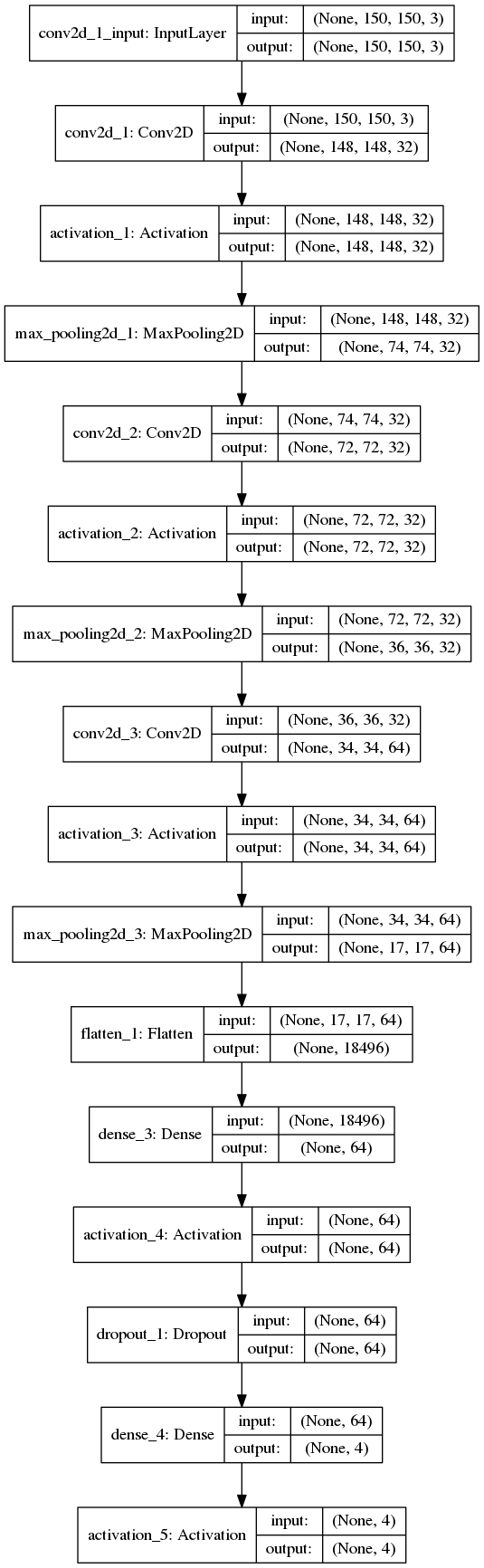}}
\caption{Deep Convolutional Neural Network used for Traffic Light recognition (Diagram made by built-in Keras function)}
\label{fig:tmodel}
\end{figure}

ETG solely uses the webcam in order to detect objects. For IARRC, a 3 layer Deep Convolution Neural Network is used as shown in figure \ref{fig:tmodel}. A 150 by 150-pixel image is formed by scaling the original 720 by 480 image down. This scaled image is put through a ConvNet. Each Convolutional layer takes a local receptive field (we used a stride length of 2) and applies random weights (which it learns over time through backpropagation) to each dimension. Each activation layer ensures that these weights are not negative. Then, a max pooling layer takes the max values within the local receptive field. This block of: Convolutional layer, Activation layer, and Max pooling layer is repeated three times. These weights are then flattened to one dimension (activation and dropout are also added to prevent negative values and overfitting, respectively) so that the CNN can predict on whether the given image contains a traffic light that is red or green as shown in figures \ref{fig:green} and \ref{fig:red}. Data is provided from Google Images.
\subsubsection{Path Planning}
As mentioned in the introduction, ETG does not use conventional path planning methods, rather a new neural network model. A DCNN is utilized for predicting steering angles while a variant of the Recurrent Neural Network, a Long Short-Term Memory System is utilized for throttle value prediction.

\subsection{Trajectory and Velocity Control}
\subsubsection{Steering Control}
\begin{figure}[ht]
\centering
\centerline{\includegraphics[width=7cm]{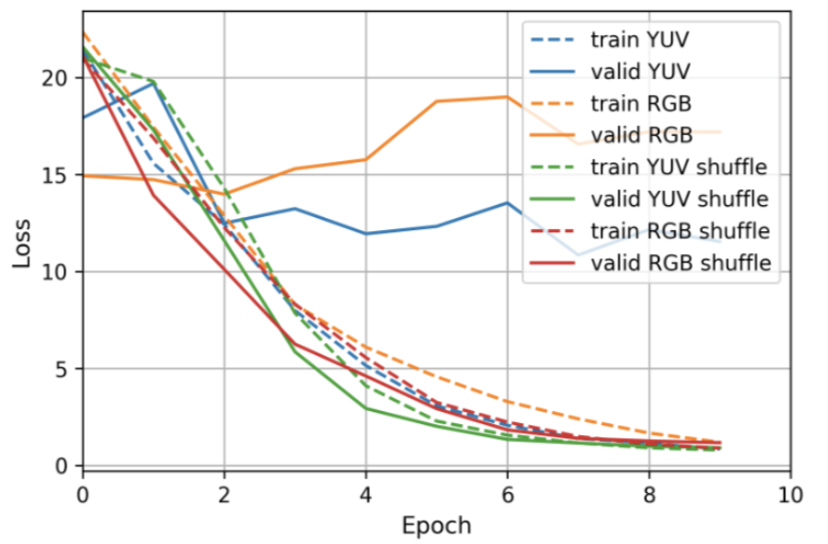}}
\caption{Loss graphs when changing colorspaces}
\label{fig:color}
\end{figure}

A Deep Convolutional Neural Network is used, similar to the network used for the object detection. The CNN as shown in figure 15 has a batch normalization and max-pooling layer of a stride length of 2 for every layer. The encoder extracts features from the first part of the DCNN before being decoded into vectors that eventually are flattened to a vector of length 10. Steering control is dependent on which of these values are turned on. If the first value is 1, then the steering is turned 100\% to the left, whereas if the last value is 1 then the steering is turned 100\% to the right. 
\newline
\indent \textbf{\textit{Training Methods}}
We utilized OpenCV and PyGame for light-weight image processing and image pre-processing. When we changed colorspaces and added a variety of tuned masks to brighten colors (such as yellow and white for the race track) we found out that our model did considerably better as shown in figure \ref{fig:color}. Unfortunately, our RPI3 ran into memory issues even when cache storages were cleared. This limited us to use our original RGB image during the entirety of the IARRC 2018.

\subsubsection{Throttle Control}
\begin{figure}[ht]
\centering
\centerline{\includegraphics[width=8cm]{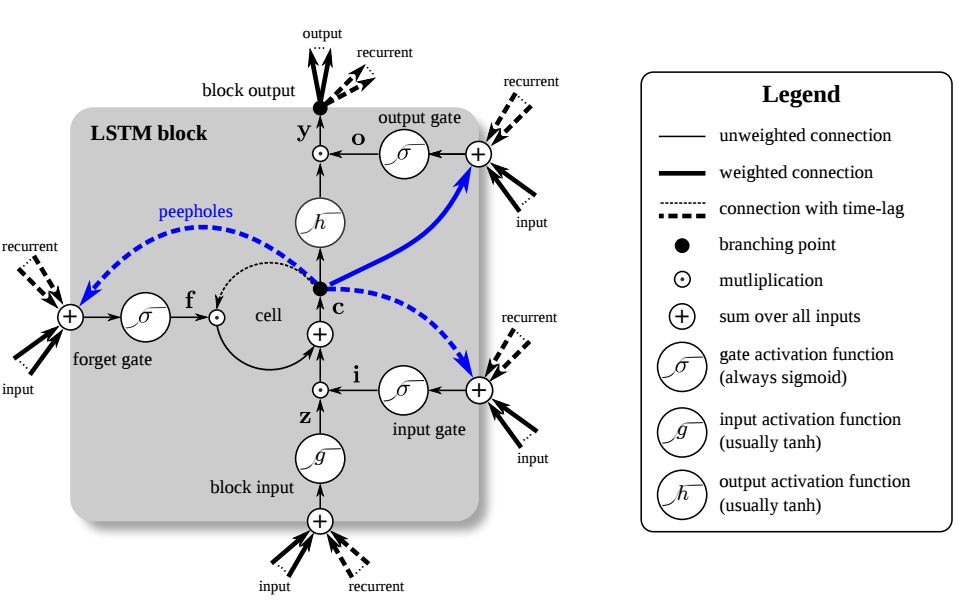}}
\caption{Representation of One LSTM [1] }
\label{fig:lstm}
\end{figure}

We found out that a Long Short-Term Memory System worked better than a DCNN through trial and error. Our LSTM system is a RNN variant that updated its weights through backpropagation (like the previous DCNN models) through the Time Distributed Layer. Our LSTM model consists of three gates within each unit (we use 2 units). 
\begin{itemize}
    \item \textbf{Input Gate:} Decides what values from input it should update
    \item \textbf{Output Gate:} Decides what values it should output based on input and memory of the unit
    \item \textbf{Forget Gate:} Decides what values it should through away from the unit
\end{itemize}
\textit{A better representation of these gates are shown in figure \ref{fig:lstm}}

To prevent overfitting we add Dropout at a value of 0.1 and a Rectified Linear Units (ReLU) layer to prevent negative values.
\newline
\indent Using a LSTM model, the throttle took the previous sequence of images and other metadata in order to produce predicted throttle values that were not drastically different from the previous time step. This meant the car sped up and slowed down with a difference of no more than 20 units in the scale of 220 (min PWM value; 0 m/s) to 420(max PWM value; 12 m/s) PWM values. Using a traditional DCNN as used before, throttle values were erratic and broke several components when testing since predictions were not based on top of one another. Predictions for both the LSTM and DCNN model (for throttle control) did have erratic values, however, the LSTM ensured that these erratic values were dropped through the Forget Gate while the DCNN used these values.
\newline
\indent \textbf{Optimization Techniques}
We used the Adam optimization algorithm, as Mohandas [4] explains, Adam is used to combine averages of previous gradients computed through backpropagation at different moments to better adaptively update weights of the model.
We also used the Mean squared error function as shown in equation \ref{eq:2}, by Hao [2], to determine the loss or accuracy between the validation and training dataset.

\begin{equation} \label{eq:2}
 \mbox{MSE} = \displaystyle\frac{1}{n}\sum_{t=1}^{n}e_t^2
\end{equation}

\begin{figure}[ht]
\centering
\hskip-1.0cm
\centerline{\includegraphics[width=11cm]{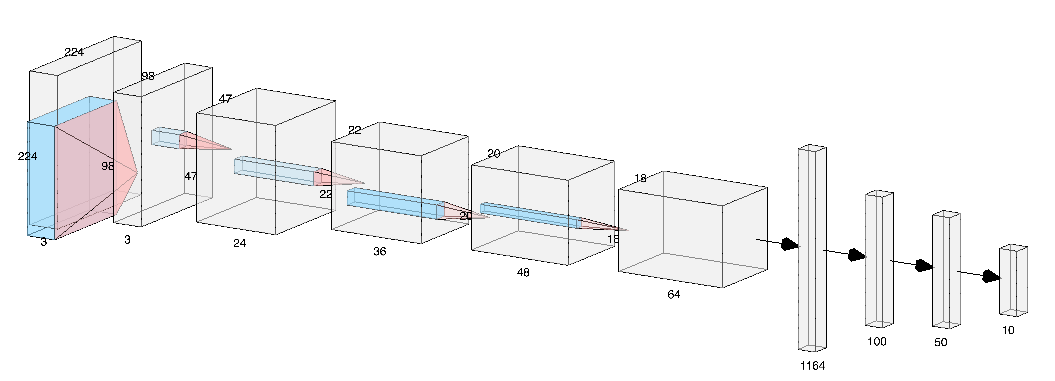}}
\caption{Deep Convolutional Neural Network for Throttle Control (Diagram made by hand)}
\label{fig:dcnn}
\end{figure}

\section{Vehicle Behavior}
\subsection{Line Detection}
\begin{figure}[ht]
\centering
\centerline{\includegraphics[width=6cm]{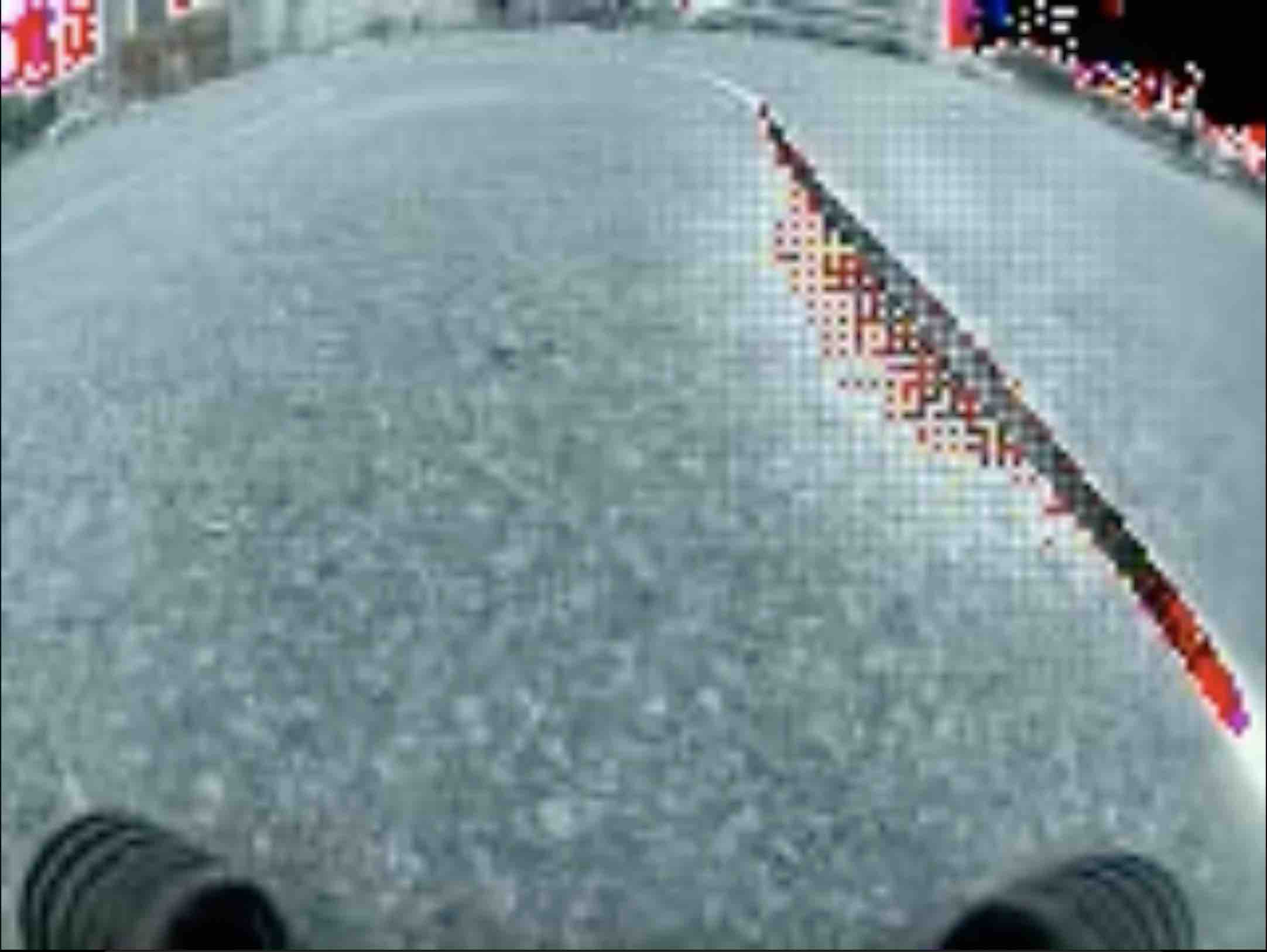}}
\caption{Line detection}
\label{fig:line}
\end{figure}

\begin{figure}[ht]
\centering
\centerline{\includegraphics[width=6cm]{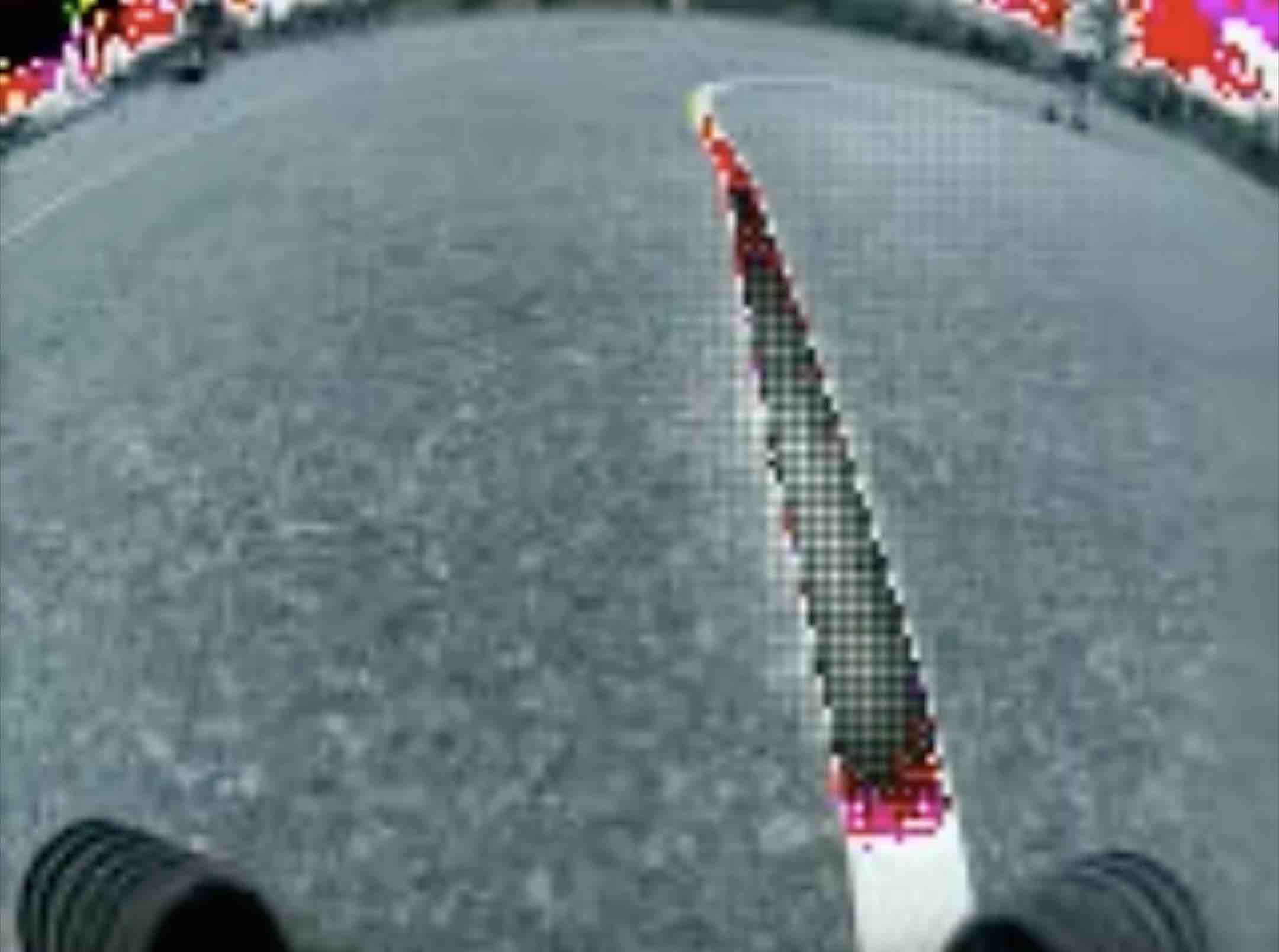}}
\caption{Clearer line detection}
\label{fig:line-clear}
\end{figure}

Without OpenCV, ETG can identify lines purely based off training data. Visualizing salient objects detected by the model, we find this out in figures \ref{fig:line} and \ref{fig:line-clear}

\subsection{Target Region using Lines}
\begin{figure}[ht]
\centering
\centerline{\includegraphics[width=6cm]{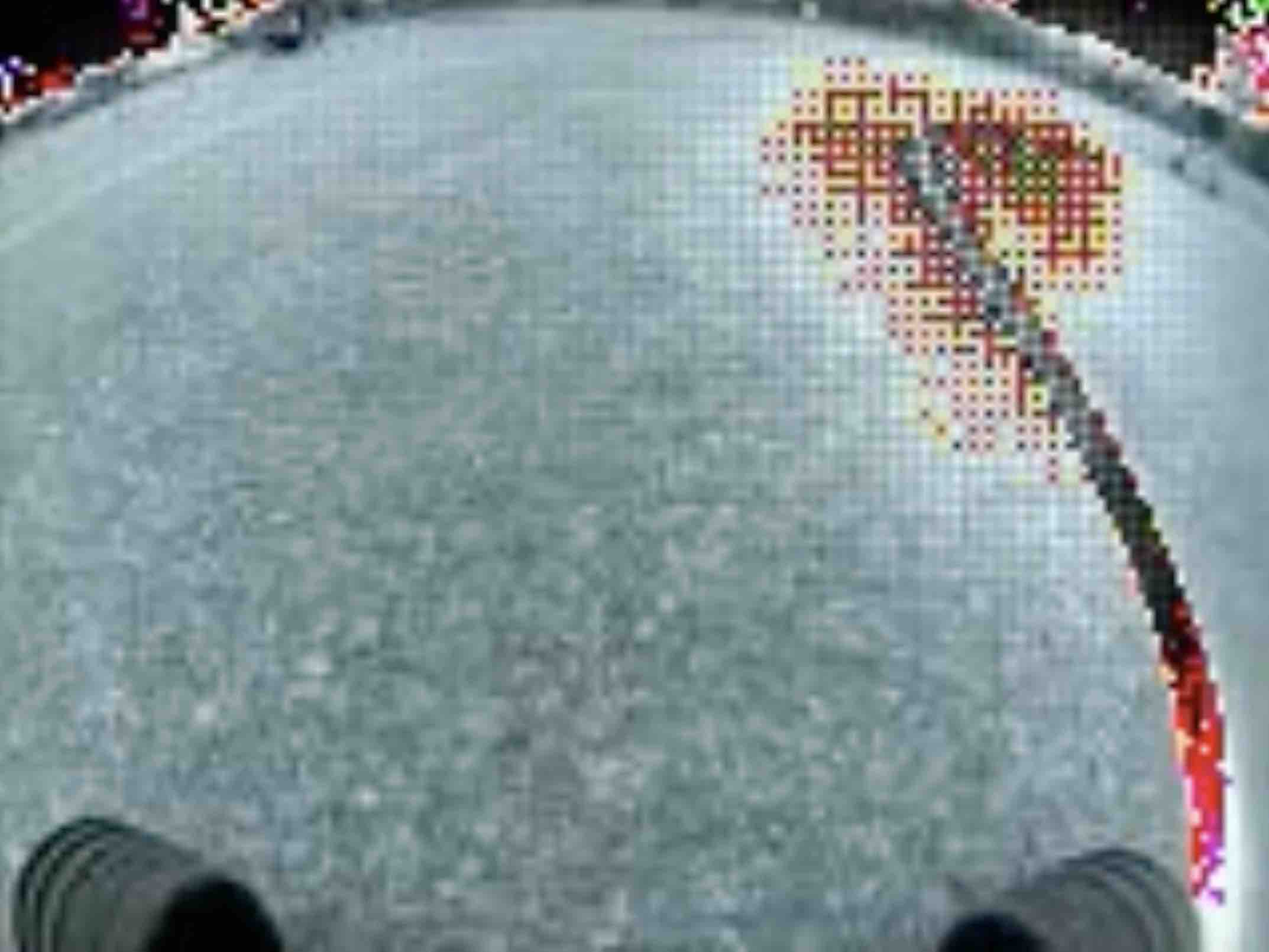}}
\caption{Line region detection}
\label{fig:line-dettect}
\end{figure}

\begin{figure}[ht]
\centering
\centerline{\includegraphics[width=6cm]{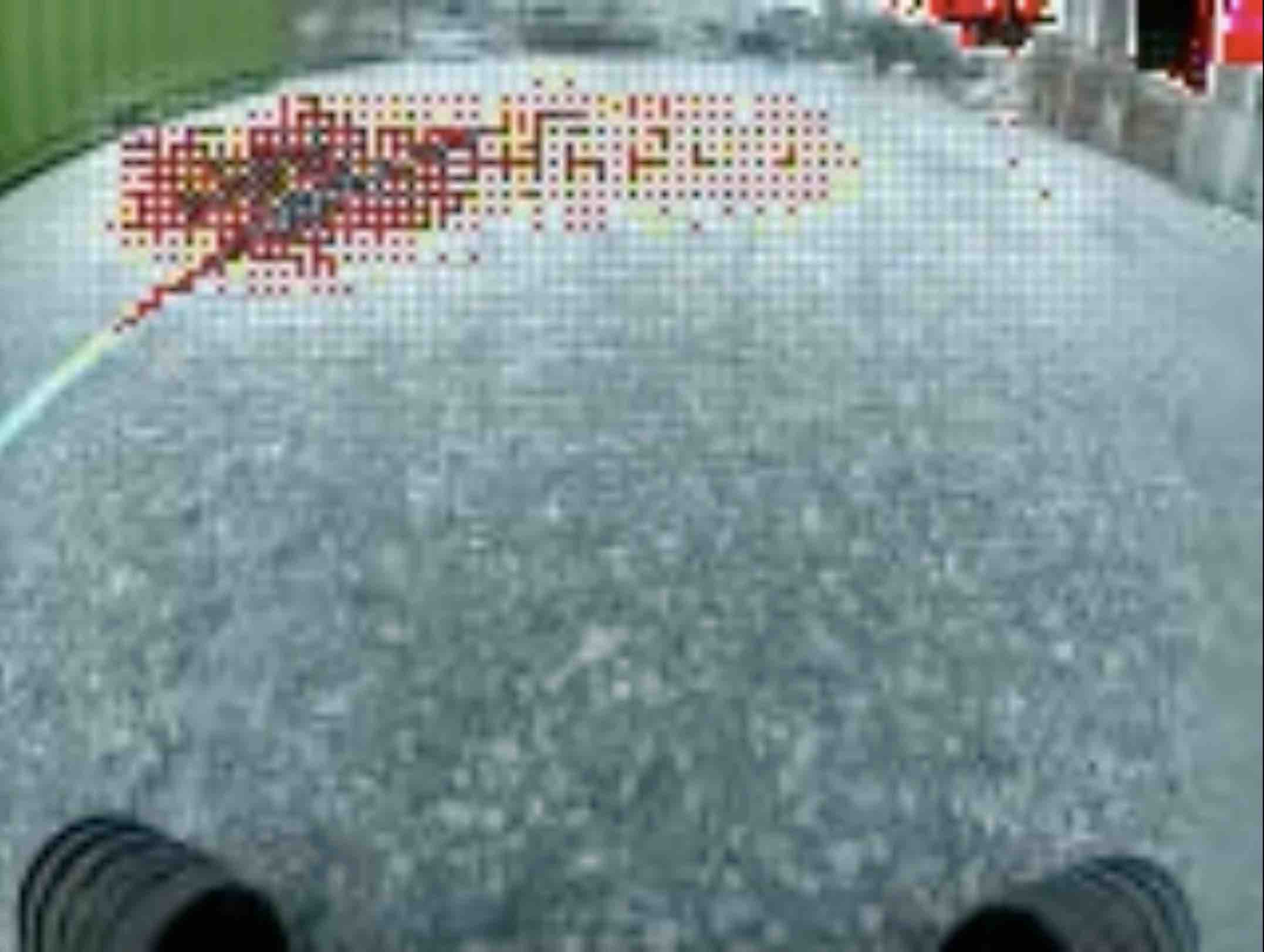}}
\caption{Line region}
\label{fig:line-region}
\end{figure}

\begin{figure}[ht]
\centering
\centerline{\includegraphics[width=6cm]{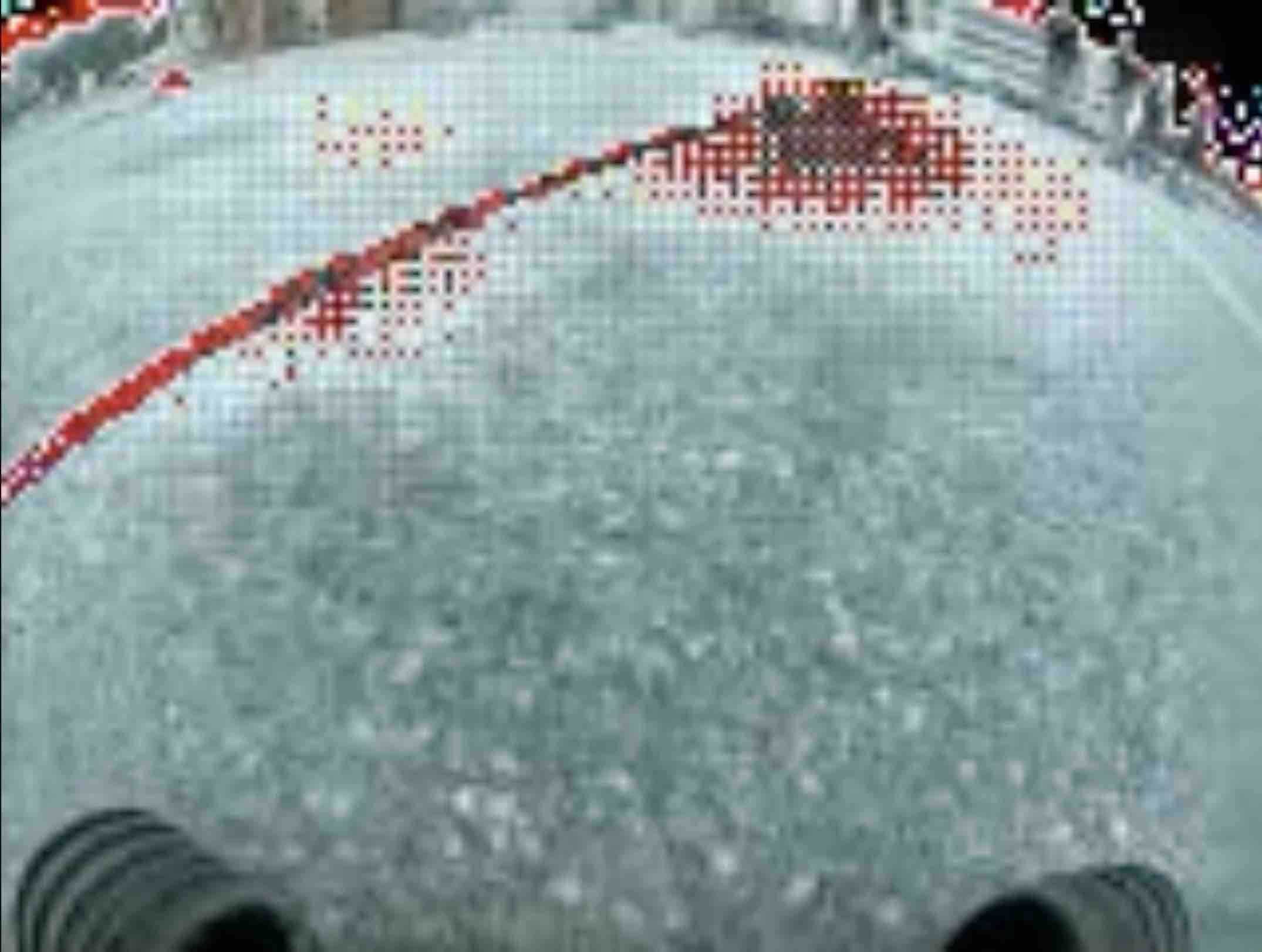}}
\caption{Line region detection}
\label{fig:line-turn}
\end{figure}

From the instance in figure \ref{fig:line-clear}, ETG starts turning left. As it is, the DCNN focuses its attention toward the right side of the viewing window, indicating that steering is predicting values that turn the vehicle toward the right as indicated in figure \ref{fig:line-dettect}. Similar behavior is seen in figures \ref{fig:line-region} \ref{fig:line-turn} where the model is trying to stay near the detected lines.

\subsection{Target Region Without using Lines}
\begin{figure}[ht]
\centering
\centerline{\includegraphics[width=6cm]{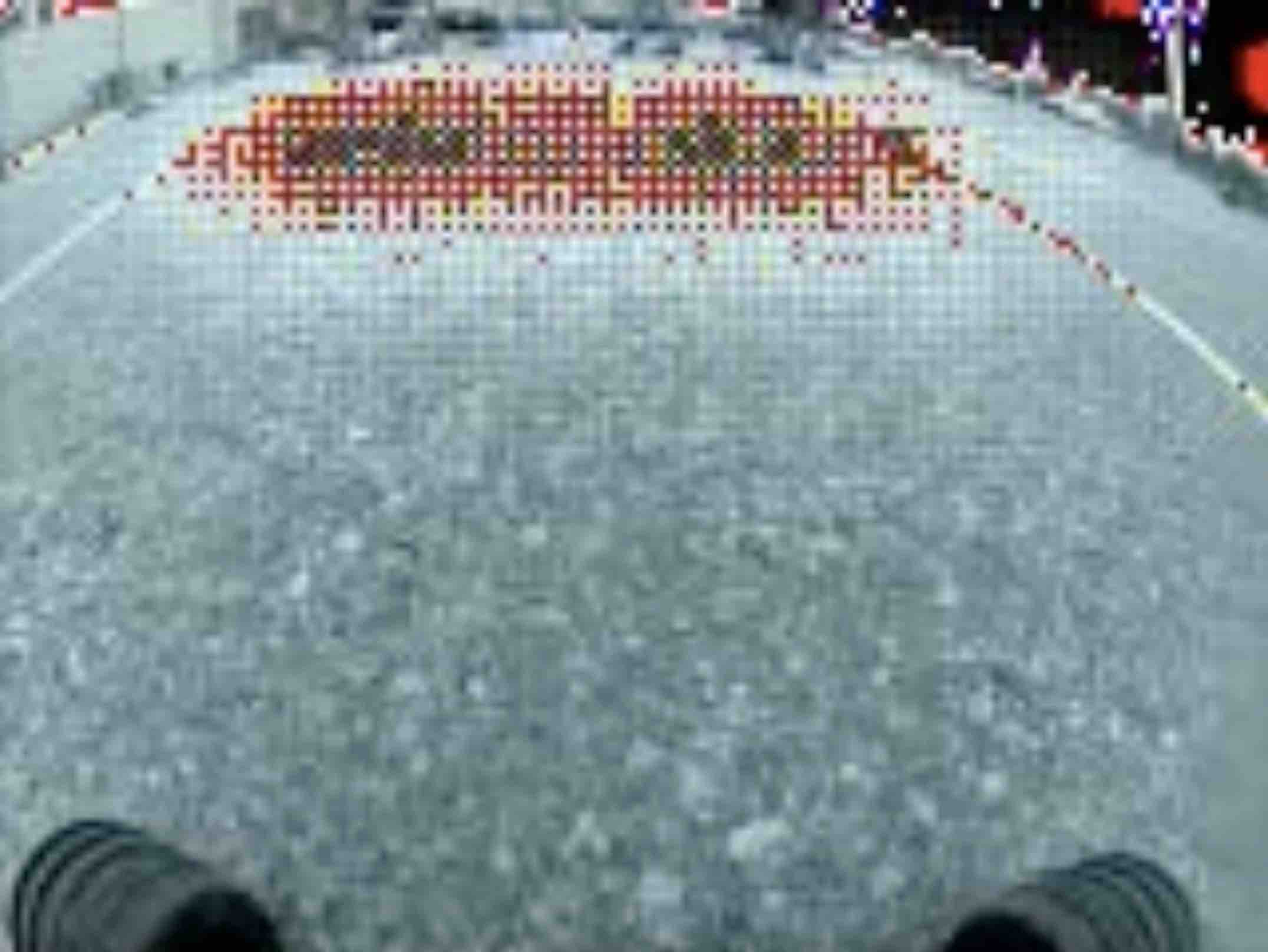}}
\caption{Line region detection}
\label{fig:no-line}
\end{figure}

\begin{figure}[ht]
\centering
\centerline{\includegraphics[width=6cm]{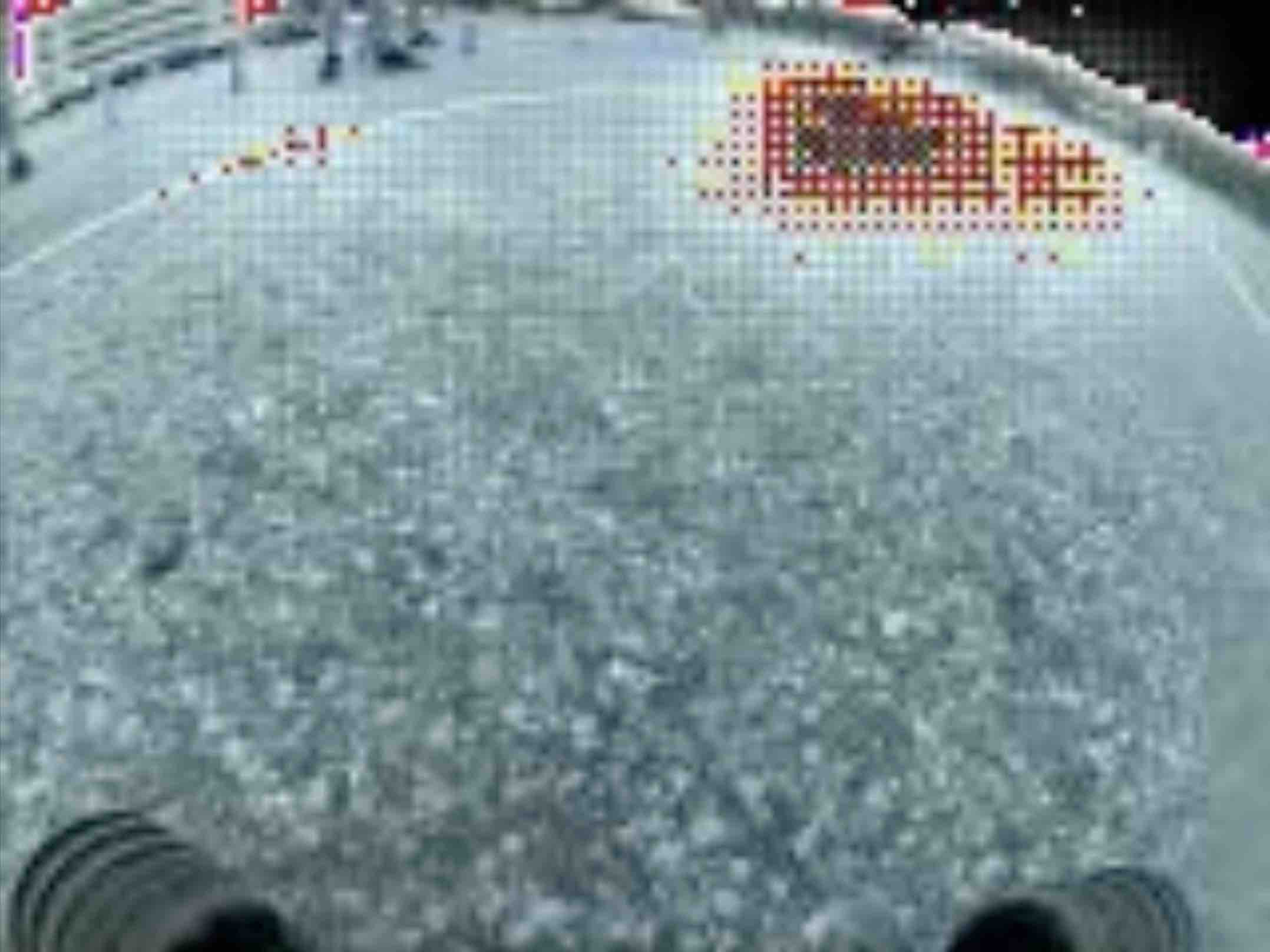}}
\caption{Line region detection}
\label{fig:no-line-2}
\end{figure}

Even without detecting the lines, the trained model (surprisingly) finds out regions of interest. This is apparent in figures \ref{fig:no-line} and \ref{fig:no-line-2} where the heatmap focuses interest on the track, not the lines themselves.

\section{Conclusion}
ETG works better than expected on courses its never seen before. As a proof of concept, it has some flaws, but proves that machine learning methods can rival traditional computationally expensive computer vision, path planning, and localization algorithms. Memory and caching issues are still a big limitation to the RPI3, which may be fixed with software updates or may be a hardware limitation. LSTM systems have proven to exceed DCNN models with throttle control and vise versa for steering control.
\subsection{Appendix}
The Monta Vista High School Team (Surya Dantuluri) competing in the 2018 International Autonomous Robot Racing Challenge are presented in Table 4:

\begin{table}[htbp]
\caption{Monta Vista High School Team}
\begin{center}
 \begin{tabular}{c c c c} 
 \hline
 \textbf{Name} & \textbf{Role} & \textbf{Hours} \\ [0.5ex]
 \hline\hline
 Surya Dantuluri & Embedded Software & 500 \\
 & Embedded Hardware & \\
 & Steering Control & \\
  & Software Integration & \\
 & Vision Systems & \\
 & Vision Systems Design & \\
 & Electro-Mechanical Design & \\
 & Machine Learning Software & \\
  & Electrical and Instrumentation & \\ [1ex] 
 \hline
  \hline
\end{tabular}
\end{center}
\end{table}

\begin{table}[htbp]
\caption{Cost Estimate}
\begin{center}
 \begin{tabular}{c c c c} 
 \hline
 Item & Actual Cost (USD) & Cost to Team \\ [0.5ex] 
 \hline\hline
 ECX 1/10 Chassis & 160.00 & 160.00 \\
 \hline
 Laptop & 1300.00 & Personal Donation \\ 
 \hline
 Raspberry Pi 3 & 40.00 & 40.00 \\
 \hline
 Arduino UNO R3 & 20.00 & 20.00\\
 \hline
 3D Printed Parts & 90.00 & 90.00 \\
 \hline
 Sensors & 50.00 & 50.00 \\
 \hline
 Portable Power Bank & 40.00 & 40.00 \\
 \hline
 5 Batteries and Charger & 300.00 & 300.00 \\
 \hline
 Motors (Brushless and Servo) & 70.00 & 70.00 \\
 \hline
 ESC(s) & 80.00 & 80.00 \\
 \hline
 Wires and Etc. & 50.00 & 50.00 \\
 \hline
 PCA 9685 & 20.00 & 20.00 \\ [1ex] 
 \hline
 \textbf{Total} & \textbf{2220.00} & \textbf{920.00}
\end{tabular}
\end{center}
\end{table}

\vspace{12pt}

\end{document}